# 1. The Exception of Humor: Iconicity, Phonemic Surprisal, Memory Recall, and Emotional Associations

Alexander Kilpatrick[1] & Maria Flaksman[2]

[1]Nagoya University of Commerce and Business
[2]Otto-Friedrich University of Bamberg

## Abstract

This meta-study investigates the relationship between humor, phonemic bigram surprisal, emotional valence, and memory recall. Previous research has demonstrated that words with higher phonemic surprisal are more easily recalled, suggesting that unpredictable sequences of phonemes enhance cognitive engagement. Another known factor that enhances recall is emotional valence whereby negative experiences and stimuli are typically recalled more readily than positive ones. This study builds upon existing research that shows that words with negative associations tend to carry more surprisal and are easier to remember. Humor—which this study shows is associated with positive emotions—is an exception as words with humorous associations both carry increased surprisal and are more memorable.

## 1 Introduction

There are a number of factors that can influence the memorability of events and stimuli. Two examples of this are probability and emotional valence where highly improbable events and stimuli are remembered with greater clarity (e.g., Ranganath & Rainer, 2003) as are those associated with negative emotions (e.g., Kensinger, 2007). This report documents a meta-study which explores these effects by examining the relationship between phonemic bigram surprisal, emotional valence, and memory recall in American English words. Specifically, it investigates whether humorous words are more surprising and memorable. The findings of this meta-study build upon existing literature by demonstrating that while negative emotional valence generally enhances memory and increases phonological surprisal, humor presents an exception. Despite its positive emotional valence, humor is associated with higher surprisal and recall accuracy. These results provide deeper insights into the interplay between phonological markedness, emotional valence, and memory retention, suggesting that humor may engage unique cognitive processes compared to other emotional categories. Important to note here that despite there being different types of humor (for a review and explanation for how deep learning models might recognize them see Chen & Soo, 2018), the present study borrows from an existing psycholinguistic experiment where participants were asked to give a high score to words that were "amusing or likely to be associated with humorous thought or language (for example, it is absurd, amusing, hilarious, playful, silly, whimsical, or laughable)" (Engelthaler & Hills, 2017). Therefore, we are exploring this one-dimensional perspective of humor.

Negativity bias, the tendency for negative information to have a greater impact on memory than positive information, plays a crucial role in shaping how we recall emotional events. Negative emotions are often remembered with greater accuracy compared to positive or neutral ones (e.g., Baumeister et al., 2001; Rozin & Royzman, 2001; LaBar & Cabeza, 2006), a phenomenon well-documented in psychology and cognitive science. This enhanced recall is primarily attributed to the evolutionary function of negative emotions which can signal potential threats. Learning to avoid negative events is more evolutionarily beneficial than engaging with positive events, leading humans to be primed for remembering and learning from negative experiences. Research by Kensinger (2007) shows that negative emotional content enhances memory retention by fostering greater attentional focus during encoding and retrieval. Additionally, studies have found that negative events are remembered more vividly due to their emotional salience, resulting in more detailed and accurate recollections (Phelps, 2004; LaBar & Cabeza, 2006). This body of research underscores the cognitive mechanisms behind the superior recall of negative emotions, emphasizing their significance in both personal memory and societal perceptions (Rozin & Royzman, 2001).

Phonemic bigram surprisal, as utilized in this study, is based on Shannon's (1948) Information Theory, which quantifies the unpredictability of events. In the context of this study, surprisal is calculated as the negative logarithm (base 2) of the probability of a particular phoneme occurring given the preceding phoneme ($P$), returning a value in bits of information. Phonemic bigram surprisal captures how unexpected a phoneme sequence is in a language. Unpredictable bigrams carry more information than predictable bigrams. Average surprisal for a word is derived by summing the information for all consecutive phoneme pairs (bigrams) and dividing by the total number of bigrams in the word.

$$Surprisal = -log_2 P \qquad (1)$$

Phonemic bigram surprisal has been shown to influence memory recall, as evidenced in a study examining iconicity and surprisal in linguistic processing (Kilpatrick & Bundgaard-Nielsen, 2024). The study demonstrated that words with high surprisal tend to be processed more slowly and less accurately but are more memorable. However, iconic words—which were already known to be easier to process and more memorable (e.g., Sidhu, Vigliocco & Pexman, 2020; Sidhu, Khachatoorian & Vigliocco, 2023)—exhibited higher average surprisal than arbitrary words. Indeed, iconic words tend to evolve towards phonemic predictability and arbitrariness over long periods of time in different stages of de-iconization (Flaksman, 2017). These stages exhibit a stochastic relationship with surprisal whereby words in early, highly iconic stages carry more surprisal than those in later, more arbitrary stages (Flaksman & Kilpatrick, In Press). This relationship between memorability, surprisal and iconicity suggests that while improbable phoneme combinations can create a cognitive disadvantage during processing, this increased effort ultimately enhances retention in long-term memory. When combined with existing research on emotion and memory, this suggests that emotional valence may be linked to phonemic surprisal. Specifically, negative words may be composed of more surprising phoneme sequences, which enhances their retention. This raises the possibility that lexemes evolve over time to become more memorable by increasing surprisal or that we are primed to create negative words with low-probability sequences. This connection between phonemic structure and emotional content provides valuable insights into the cognitive mechanisms underlying memory retention.

Phonemic surprisal could serve as an important additional datapoint in machine learning algorithms focused on emotion and sentiment analysis. By quantifying the predictability of phoneme sequences, phonemic surprisal offers insights into how sound patterns might relate to emotional valence in language. Current sentiment analysis algorithms typically rely on lexical and syntactic features, often overlooking the phonological aspects that could improve model accuracy. This position aligns with earlier research (Kilpatrick, 2023) that explored the utility of iconic associations between phonemes and various emotions and sentiments. In that study, researchers constructed machine learning algorithms to predict emotions and sentiments using the phonemes in each word. Model feature importance scores revealed that, while it was difficult to distinguish fine-grained emotional differences, general positivity and negativity was stochastically reflected in phonemes. Interestingly, the algorithms constructed to predict negative emotions (Anger, Disgust, Fear, Negative Valence, and Sadness) performed better than those constructed to predict positive emotions (Anticipation, Joy, Positive Valence, Surprise, and Trust), suggesting that iconic negative associations are more robustly reflected than positive associations.

Increased surprisal in iconic words represents a form of phonological markedness that goes hand in hand with other observed iconic markedness strategies (Voeltz & Kilian-Hatz, 2001) including the use of phonotactic violations (e.g., *vroom* [vɹum]), non-native speech sounds (e.g., *ugh* [əx]), gemination (e.g., GRRRR! [gr:]), or vowel lengthening (e.g., WHAAT? [wæ:t]). Dingemanse and Thompson (2020) explore the relationship between humor and markedness through the lens of iconicity. They propose that structural markedness is a key factor underlying perceptions of both funniness and iconicity in words. Marked cues function as metacommunicative signals, drawing attention to words as playful and performative. This research suggests that playful and poetic elements are integral to the lexicon, highlighting the intersection of humor and markedness in language. In the context of the present study, this suggests that words with humorous associations should carry more information than those without. Imitative words—particularly ideophones—are expressive and more likely to violate phonological, morphological and syntactic norms (Dingemanse, 2017 Dingemanse & Akita, 2017). As both humorous and iconic words are related to expressivity, this relationship is worth further investigation which we attempt in the present study. For example, we expect words like *booby* ($M$ = 4.07), *waddle* ($M$ = 4.05) or *gaggle* ($M$ = 3.82) to have higher average surprisal than words like *torture* ($M$ = 1.26), war ($M$ = 1.34), or *casket* ($M$ = 1.38), where numbers in parentheses represent humor Likert averages. However, this prediction is at odds with our earlier prediction that words with negative associations should carry more information because humor is typically associated with positive valence. This study seeks to reconcile these seemingly contradictory predictions.

This study builds on prior research investigating the relationship between phonemic surprisal, emotional valence, and memory, with a particular focus on how humor functions within this framework. While previous findings have highlighted the role of negativity bias in enhancing memory recall (Kilpatrick, Under Review), humor represents a unique case of a positive emotional valence associated with high surprise. By examining words rated for their humor, we aim to determine whether the cognitive mechanisms that enhance memorability in negative valence also apply to humor, and how these effects differ between iconic and non-iconic words.

## 2 Method

Data here: https://shorturl.at/2SXvO. Phonemic bigram surprisal was calculated by cross-referencing the SUBLEX-US corpus (Brysbaert & New, 2009) with the CMU Pronouncing Dictionary (Weide, 1999) to obtain phonemic transcriptions and frequencies. A more detailed explanation of this process is provided in the above link. This combined dataset was then cross-referenced with existing datasets to provide morpheme counts (Sánchez-Gutiérrez, 2018) and parts of speech (Brysbaert, New, & Keuleers, 2012) because number of morphemes and word classification have been shown to influence surprisal (Kilpatrick & Bundgaard-Nielsen, 2024). Iconicity ratings were obtained from an existing experiment (Winter et al., 2023) where American English speakers were asked to provide Likert scale ratings to words according to how much they "sound like" their meaning. The memory recall data comes from a pre-existing psycholinguistic experiment (Cortese, Khanna, & Hacker, 2010) which involved the training of 120 undergraduate students on a list of words in one experimental session and the testing of their recall accuracy in a second session within the same week.

This study draws from three existing experiments for the emotion data. Firstly, there are ten emotions—Anger, Anticipation, Disgust, Fear, Joy, Negative, Positive, Sadness, Surprise, and Trust—taken from the NRC Emotion Lexicon (Mohammad & Turney, 2013) where American English-speaking participants were asked to provide binary responses to words according to whether they associate each word with a particular emotion. The NRC_Valence (Mohammad & Turney, 2013) variable also comes from the NRC emotion lexicon while G_Valence (Scott et al., 2019) comes from the Glasgow Norms which was collected from English speaking participants in Scotland and is included to explore potential crossover into other variants of English. Lastly, the Humor variable comes from an online study (Engelthaler & Hills, 2018) where English-speaking participants were asked to rate how humorous words are on a 5-point Likert scale. In that study, participants were asked to rank words where at one end of the scale, words are "dull or unfunny" and at the other, "absurd, amusing, hilarious, playful, silly, whimsical, or laughable" (Engelthaler & Hills, 2018). The original study made no distinction between different types of humor and there is no way to disentangle differences between, irony, sarcasm, dark humor, or wordplay. No samples were excluded from the models except in the case of missing data. That is, words like *glimmer*, *whisper*, and *crunch*, are iconic, but are not particularly humorous nor are they seemingly associated with emotional valence. Despite not carrying said associations, they were included in the following analyses.

The emotional response variables are measured using three distinct types of scales. For emotions from the NRC emotion lexicon, such as Fear, responses are measured on a binary scale, where participants rate the presence or absence of a single emotion from 0 (neutral) to 1 (fearful). On the other hand, the humor dataset presents a one-tailed continuous scale represented by averages of Likert responses from 1 (neutral) to 5 (humorous). Lastly, the valence variables are assessed on a two-tailed scale, where ratings range from 0 (negative) to 7 (positive), with 3.5 representing a neutral emotional state. This scale accounts for both positive and negative valence, capturing a bidirectional emotional response. That noted, there is undoubtedly some measure of bidirectionality in the binary variables, particularly the Negative and Positive variables from the NRC emotion lexicon.

This data is used in two series of multiple linear regression models. The first series is designed to explore the relationship between emotions—which are included as the dependent variables—and average bigram surprisal (Average_Surprisal) which is included alongside iconicity ratings (Iconicity_Rating), phonemic length (Phoneme_Length), morphemic length (Morpheme_Length), and parts of speech (PoS) categories. Here, we predict that those emotions associated with positivity (e.g., Anticipation, Joy, Positive, Surprisal, and Trust) will carry less information than those associated with negativity (e.g., Anger, Disgust, Fear, Negative, and Sadness). We predict that this pattern will be exhibited more robustly in the Valence variables due to their bi-directional nature. Humor--or at least words assigned high humor ratings in Engelthaler & Hills, (2018)—is predicted to carry more information despite being typically associated with positivity. In the second series of models, the emotion variables are included as independent variables, and the results of the memory recall experiment are included as the dependent variables. Here, we predict that same pattern whereby negative emotions will exhibit a positive correlation with memory recall, even when average bigram surprisal is taken into consideration. Again, we expect Humor to buck this trend and exhibit an increased memory recall accuracy.

## 3 Results

Firstly, to test the assumption that humor has a generally positive association, we ran two simple linear regression analyses, using Humor ratings (Engelthaler & Hills, 2018) as the dependent variable and valence ratings (NRC_Valence and G_Valence) as the predictor variables. In both models, valence was a significant predictor of humor, indicating a positive correlation between Humor and positive valence ($p < 0.001$ in both models).

| Variable | G_Valence | NRC_Valence |
|---|---|---|
| (Intercept) | 28.03*** | 37.22*** |
| **Average_Surprisal** | **-2.11*** | **-4.51*** |
| **Iconicity_Rating** | **-5.32*** | **-10.31*** |
| Phoneme_Length | 0.37 | -0.18 |
| Morpheme_Length | 1.31 | -1.24 |
| PoS_Adverb | 1.61 | 2.89** |
| PoS_Determiner | 0.41 | 0.029 |
| PoS_Interjection | -0.05 | 0.87 |
| PoS_Name | -0.18 | 2.40* |
| PoS_Noun | 1.70 | 5.37*** |
| PoS_Number | 1.337 | 0.904 |
| PoS_Preposition | -0.22 | 0.45 |
| PoS_Pronoun | 0.45 | 1.54 |
| PoS_Unclassified | 0.28 | 0.14 |
| PoS_Verb | -1.48 | 0.74 |

Table 1: Results of the two valence models. Asterisks denote significance (* $p < 0.05$, ** $p < 0.01$, *** $p < 0.001$).

Two models were constructed using the two-tailed valence variables from the NRC emotion lexicon and the Glasgow norms (Table 1). Both exhibited a significant negative effect of both average surprisal and iconicity ratings revealing that negative valence is associated with both increased surprisal and iconicity. In other words, words associated with negative valence are made up of more unpredictable bigrams and negative valence is more robustly expressed than positive valence in iconic associations.

Following this, a series of multiple linear regression models were constructed using the one-tailed emotion variables from the NRC emotion lexicon (Table 2). These models show a general trend whereby negative emotions carry more average surprisal. Important to note here is that this was only significant with Disgust, which exhibited a significant positive correlation, and Anticipation and Joy which exhibited significant negative correlations. Almost all variables demonstrated a positive correlation with iconicity ratings except for Anticipation which was not significant and Positive which revealed a significant negative relationship with iconicity ratings.

|  | Negative Emotions | | | | | Positive Emotions | | | | |
| --- | --- | --- | --- | --- | --- | --- | --- | --- | --- | --- |
|  | Anger | Disgust | Fear | Negative | Sadness | Anticipation | Joy | Positive | Surprise | Trust |
| (Intercept) | -1.305 | 0.043 | -0.909 | 0.251 | 0.447 | 3.298*** | 2.718** | 7.781*** | -2.732** | 6.267*** |
| **Average_Surprisal** | **0.247** | **2.574*** | **-0.69** | **1.543** | **0.085** | **-3.029**** | **-1.989*** | **-1.926** | **0.902** | **-1.523** |
| **Iconicity_Rating** | **6.495**** | **5.499**** | **7.169**** | **9.963**** | **4.062**** | **0.804** | **2.749**** | **-3.43**** | **6.431**** | **-5.886**** |
| Phoneme_Length | 3.181** | 1.416 | 2.593** | 2.911** | 1.057 | 1.277 | 1.679. | 4.395*** | 3.436*** | 2.854** |
| Morpheme_Length | 0.685 | 0.347 | 0.611 | 1.847. | 3.142** | 0.668 | -0.689 | 0.481 | -1.399 | 0.179 |
| PoS_Adverb | -2.606** | -2.539* | -2.148* | -3.869*** | -1.294 | -0.019 | -0.759 | -1.353 | 1.132 | -0.989 |
| PoS_Determiner | -0.301 | -0.441 | -0.312 | -0.696 | -0.335 | -0.23 | -0.306 | -0.533 | -0.163 | -0.255 |
| PoS_Interjection | -1.391 | -0.893 | -1.393 | -0.197 | -1.367 | 0.586 | -1.012 | 0.149 | 0.469 | -0.384 |
| PoS_Name | -1.45 | -2.683** | -0.18 | -3.909*** | -1.657. | -0.553 | -0.442 | -1.651 | -1.444 | 0.257 |
| PoS_Noun | -2.205* | -7.132*** | 0.745 | -9.067*** | -3.462*** | 0.378 | -3.275** | -7.027*** | 0.176 | 0.556 |
| PoS_Number | -1.163 | -1.621 | -1.15 | -2.623** | -1.398 | -0.918 | -1.1 | -2.278* | -0.54 | -1.213 |
| PoS_Preposition | 0.798 | -1.517 | -1.112 | -1.118 | -1.288 | -0.881 | -1.065 | -0.708 | -0.577 | -1.192 |
| PoS_Pronoun | -0.414 | -0.554 | -0.431 | -0.917 | -0.455 | -0.39 | -0.469 | -0.879 | -0.212 | -0.477 |
| PoS_Unclassified | -0.55 | -0.792 | 1.645 | -1.212 | -0.555 | 2.488* | -0.433 | 1.228 | -0.368 | -0.188 |
| PoS_Verb | 1.704. | -6.244*** | -0.269 | -3.253** | -1.014 | 2.619** | -1.759. | -4.875*** | 1.616 | 1.242 |

Table 2: Results of the models run using the one-tailed NRC emotion variables. Asterisks denote significance (* $p < 0.05$, ** $p < 0.01$, *** $p < 0.001$).

| **Variable** | **Humor** |
| --- | --- |
| (Intercept) | 38.202*** |
| **Average_Surprisal** | 3.125** |
| **Iconicity_Rating** | 18.006*** |
| Phoneme_Length | -1.368 |
| Morpheme_Length | -6.374*** |
| PoS_Adverb | -0.813 |
| PoS_Interjection | 1.497 |
| PoS_Name | -0.768 |
| PoS_Noun | 2.449* |
| PoS_Number | -1.823 |
| PoS_Preposition | 0.593 |
| PoS_Verb | -1.386 |

Table 3: Results of the humor model. Asterisks denote significance (* $p < 0.05$, ** $p < 0.01$, *** $p < 0.001$).

| **Variable** | **G_Valence** | **NRC_Valence** |
| --- | --- | --- |
| (Intercept) | 29.581*** | 15.168*** |
| **Valence** | **-1.009** | **-5.154**** |
| **Average_Surprisal** | **4.899**** | **7.098**** |
| **Iconicity_Rating** | 1.707 | 2.582** |
| Phoneme_Length | -2.833** | -2.645** |
| Morpheme_Length | -2.345* | -3.637*** |
| PoS_Adverb | -0.347 | -0.973 |
| PoS_Interjection |  | 0.371 |
| PoS_Name | 2.581** | 2.64** |
| PoS_Noun | 6.666*** | 5.657*** |
| PoS_Number | -1.774 | 0.481 |
| PoS_Preposition | -0.936 | -1.034 |
| PoS_Verb | -7.083*** | -10.398*** |

Table 4: Results of the two memory/valence models. Asterisks denote significance (* $p < 0.05$, ** $p < 0.01$, *** $p < 0.001$).

Humor was also tested as a dependent variable (Table 3). Despite being associated with positive valence, it exhibited a significant positive relationship with average surprisal. Humor was also found to be a significant predictor of iconicity ratings where high humor ratings correlated with high iconicity. In another way, words associated with humor like *oomph* (Humor = 3.93; Average Surprisal = 6.26; Iconicity = 6.92) are made up of unpredictable bigrams and are iconic while words that are not associated with humor like *cancer* (Humor = 1.46; Average Surprisal = 3.62; Iconicity = 2.83) are less surprising and less iconic.

All models thus far were then reconstructed except the emotions were included as an independent variable and the dependent variable for each model was the memory recall accuracy results. First, we reconstructed the valence models (Table 4) and found that negative valence was associated with improved memory recall; however, this relationship was only significant in the NRC emotion lexicon model ($p < 0.001$). For both models, increased average surprisal was associated with increased memory recall.

The pattern between negative valence and memory recall was also exhibited in the models constructed using the NRC emotion lexicon emotions (Table 5). Here, Disgust, Fear, and Negative were significant predictors of increased memory recall while a significant negative correlation was observed between positive emotions, Anticipation, Positive, and Surprise. In all models, high average surprisal was a significant predictor of high memory accuracy.

|  | Negative Emotions | | | | | Positive Emotions | | | | |
|---|---|---|---|---|---|---|---|---|---|---|
|  | Anger | Disgust | Fear | Negative | Sadness | Anticipation | Joy | Positive | Surprise | Trust |
| (Intercept) | 29.556*** | 29.71*** | 29.569*** | 29.511*** | 29.511*** | 29.662*** | 29.442*** | 29.697*** | 29.515*** | 29.444*** |
| **Emotion** | **1.348** | **5.729*** | **1.392** | **3.346*** | **2.074*** | **-2.787*** | **0.887** | **-2.842*** | **-2.282*** | **-0.578** |
| **Average_Surprisal** | **5.495*** | **5.191*** | **5.464*** | **5.377*** | **5.503*** | **5.436*** | **5.505*** | **5.489*** | **5.526*** | **5.485*** |
| **Iconicity_Rating** | **2.098*** | **1.832.** | **2.083*** | **1.71.** | **2.107*** | **2.154*** | **2.214*** | **2.017*** | **2.357*** | **2.131*** |
| Phoneme_Length | -3.801*** | -3.946*** | -3.832*** | -3.926*** | -3.812*** | -3.753*** | -3.754*** | -3.631*** | -3.656*** | -3.734*** |
| Morpheme_Length | -4.209*** | -4.108*** | -4.199*** | -4.149*** | -4.214*** | -4.19*** | -4.197*** | -4.247*** | -4.293*** | -4.228*** |
| PoS_Adverb | -1.699. | -1.661. | -1.7. | -1.657. | -1.688. | -1.607 | -1.741. | -1.665. | -1.719. | -1.689. |
| PoS_Interjection | 0.251 | 0.305 | 0.252 | 0.174 | 0.259 | 0.23 | 0.245 | 0.221 | 0.394 | 0.241 |
| PoS_Name | 2.084* | 2.162* | 2.027* | 2.165* | 2.133* | 2.078* | 2.084* | 2.002* | 2.095* | 2.077* |
| PoS_Noun | 2.15* | 2.555* | 2.121* | 2.467* | 2.258* | 2.09* | 2.174* | 1.895. | 2.09* | 2.133* |
| PoS_Number | 0.009 | 0.063 | 0.009 | 0.056 | 0.021 | -0.018 | 0.009 | -0.045 | -0.01 | -0.004 |
| PoS_Verb | -10.183 | -9.798*** | -10.165 | -9.975*** | -10.092*** | -10.087*** | -10.146*** | -10.297*** | -10.14*** | -10.136 |

Table 5: Results of the models run using the memory recall results and the one-tailed NRC emotion variables. Asterisks denote significance (* p < 0.05, ** p < 0.01, *** p < 0.001).

Lastly, the humor model was reconstructed (Table 6). It revealed that humor follows the same pattern as negative emotions, where words associated with humor are recalled with greater accuracy than humorless words.

| **Variable** | **Humor** |
|---|---|
| (Intercept) | 15.168*** |
| **Humor** | **17.628*** |
| **Average_Surprisal** | **5.371*** |
| **Iconicity_Rating** | **-4.625*** |
| Phoneme_Length | -2.203* |
| Morpheme_Length | -1.402 |
| PoS_Adverb | -1.949. |
| PoS_Interjection | -0.053 |
| PoS_Name | 1.872. |
| PoS_Noun | 3.563*** |
| PoS_Number | 1.474 |
| PoS_Preposition | -1.437 |
| PoS_Verb | -5.353*** |

Table 6: Results of the humor model. Asterisks denote significance (* p < 0.05, ** p < 0.01, *** p < 0.001).

## 4 Discussion

In this study, we explore the hypothesis that humor may be linked to high phonological surprisal and improved memory recall. Evidence from other studies suggests that negative valence is associated with high surprisal (Kilpatrick, Under Review), and that surprisal plays a role in cognitive processing. For instance, Kilpatrick and Bundgaard-Nielsen (2024) found that words with higher surprisal are more difficult to process by also more likely to be recalled in memory tasks. Building on this, we investigate whether the increased surprisal of negatively valanced words might similarly influence cognitive mechanisms, such as memory and attention. We also explored how these linguistic features influence and interact with humor, revealing that words with humorous associations perform like negatively valanced words despite being stochastically positive.

The Negativity bias reflects the human tendency to prioritize and more deeply process negative stimuli over neutral or positive ones. Evolutionarily, this bias likely developed because organisms that were more attuned to dangers or threats were better equipped to avoid harm, thus improving their chances of survival (Baumeister et al., 2001). As a result, negative stimuli tend to attract more cognitive resources, leading to enhanced memory, attention, and decision-making processes. The negativity bias seems to be encoded into words via transitional probability—here, measured by surprisal—with more surprising or less predictable words often carrying negative emotional valence. This suggests that high-surprisal words may serve as cognitive markers, alerting individuals to potentially dangerous or emotionally charged situations, thus reinforcing the prioritization of such information in human cognition. For example, Disgust, an emotion deeply rooted in survival, due to its connection with avoidance of harmful substances or contamination, was found to be highly memorable and surprising. Phonological cues in disgust-related words might serve as a cognitive alarm, triggering a rapid response to potential dangers, such as toxic or unsafe materials. This adaptive mechanism ensures that words linked to disgust demand attention and are memorable, reinforcing their role in threat detection and avoidance. Further testing of this hypothesis should explore how words with dangerous associations—rather than negative valence—are reflected in phonemic surprisal and their effects on memory retention.

However, our results show that humor follows the same patterns as negative stimuli, suggesting that humor may exploit similar cognitive mechanisms. Just as negative stimuli demand attention and leave a lasting impression, humor, which often subverts expectations or highlights absurdities, may trigger heightened cognitive engagement through surprise or incongruity. While both humor and negativity utilize elements of unpredictability, humor diverges in its social function. Suls's Two-Stage Model (1972) for the Appreciation of Jokes provides a useful framework for understanding this. In the first stage, the punch line of a joke is surprising, incongruous, and may be perceived as threatening due to its unexpected nature, eliciting a response akin to the cognitive engagement seen with negative stimuli. In the second stage, the listener resolves the incongruity, leading to a sense of relief and the recognition of humor, which ultimately results in a positive emotional response. This aligns with our findings that humor, while engaging cognitive processes similar to those of negative stimuli, also embodies a transition from initial surprise to positive social outcomes, such as connection and bonding. Unlike negative stimuli, which may trigger responses tied to alertness or threat, humor's playful disruption fosters a social and positive emotional environment.

Dingemanse and Thompson's (2020) work further illuminates these dynamics, proposing that structural markedness, including phonological markedness, underpins perceptions of both humor and iconicity. This aligns with our findings that humor and negative valence share cognitive engagement mechanisms, suggesting that both types of words are marked in ways that enhance memorability. Playful and performative elements like increased surprisal or other markedness strategies inherent in humorous language may interact with the cognitive mechanisms associated with negativity bias, drawing attention to their phonological structure and making them more memorable. Surprisal serves as an objective measure of phonological markedness by quantifying the predictability of linguistic units based on their transitional probabilities within a given context. Unlike subjective judgments that can vary among speakers or listeners, surprisal provides a statistical framework for assessing the complexity of markedness in iconic words. Words or sounds that exhibit higher surprisal values are often those that are less predictable, indicating a higher degree of markedness. This objective measure allows researchers to analyze the cognitive processing of language without relying on potentially biased human assessments.

Future research should investigate languages other than American English which might reveal cross-linguistic patterns. If the patterns observed in this study are observed in other languages, then this would suggest that they are innate, rather than cultural, further enhancing our understanding of the interplay between language, emotion, and cognitive processing. Another direction of research is differentiating different types of humor: (1) humor based on use of highly colloquial language highly saturated with iconic words (as words from the original study of Engelthaler & Hills (2018), (2) farcial humor or comedy of situation (e.g., Shakespeare's *Much Ado about Nothing*) where ridiculous dramatic situations fall into category of "highly improbable events" discussed in Ranganath & Rainer (2003), or (3) dark humor and sarcasm (which is a mixture of negativity (see above) and low probability, both of which should, theoretically, increase memorability.

In conclusion, our findings suggest that negativity bias is reflected in the phonological surprisal of words, with negatively valenced words comprising more improbable sequences of sounds and demonstrating greater memorability. Interestingly, humor subverts this trend, exhibiting both increased surprisal and memorability, highlighting its unique role in language.

## Acknowledgments

We wish to thank the researchers who made their data publicly available so that we could conduct this analysis. This project was funded by the Japan Society for the Promotion of Science (# 20K13055).